\documentclass[journal]{IEEEtran}

\usepackage[noend]{algpseudocode}

\usepackage{cite}
\usepackage{url}

\usepackage{textcomp}
\usepackage{xcolor}
\usepackage{hyperref}

\usepackage{stfloats}   

\usepackage{booktabs}	
\usepackage{diagbox}    
\usepackage{multirow}   

\usepackage{verbatim}	

\usepackage{amsmath,amssymb,amsfonts,graphicx}
\usepackage{epstopdf}

\usepackage[ruled,linesnumbered]{algorithm2e}

\def\BibTeX{{\rm B\kern-.05em{\sc i\kern-.025em b}\kern-.08em
		T\kern-.1667em\lower.7ex\hbox{E}\kern-.125emX}}

\usepackage{amssymb}
\usepackage{hyperref}

\usepackage[switch]{lineno}   

\begin{document}

\title{MDL-based Compressing Sequential Rules}

\author{Xinhong Chen, Wensheng Gan*,~\IEEEmembership{Member,~IEEE,}
	Shicheng Wan, and Tianlong Gu
	
\thanks{This research was supported in part by the National Natural Science Foundation of China (Nos. 62272196 and 62002136), Natural Science Foundation of Guangdong Province (No. 2022A1515011861), Guangzhou Basic and Applied Basic Research Foundation (No. 202102020277), and the Young Scholar Program of Pazhou Lab (No. PZL2021KF0023). (Corresponding author: Wensheng Gan)}
	
	\thanks{Xinhong Chen and Tianlong Gu are with the College of Cyber Security, Jinan University, Guangzhou 510632, China (E-mail: xinhchen00@gmail.com, gutianlong@jnu.edu.cn)}
	
	\thanks{Wensheng Gan is with the College of Cyber Security, Jinan University, Guangzhou 510632, China; and also with Pazhou Lab, Guangzhou 510330, China (E-mail: wsgan001@gmail.com)}
	
	\thanks{Shicheng Wan is with the Department of Computer Sciences, Guangdong University of Technology, Guangzhou 510006, China. (E-mail: scwan1998@gmail.com)} 

}

\maketitle

\begin{abstract}	
	Nowadays, with the rapid development of the Internet, the era of big data has come. The Internet generates huge amounts of data every day. However, extracting meaningful information from massive data is like looking for a needle in a haystack. Data mining techniques can provide various feasible methods to solve this problem. At present, many sequential rule mining (SRM) algorithms are presented to find sequential rules in databases with sequential characteristics. These rules help people extract a lot of meaningful information from massive amounts of data. How can we achieve compression of mined results and reduce data size to save storage space and transmission time? Until now, there has been little research on the compression of SRM. In this paper, combined with the Minimum Description Length (MDL) principle and under the two metrics (\textit{support} and \textit{confidence}), we introduce the problem of compression of SRM and also propose a solution named ComSR for MDL-based compressing of sequential rules based on the designed sequential rule coding scheme. To our knowledge, we are the first to use sequential rules to encode an entire database. A heuristic method is proposed to find a set of compact and meaningful sequential rules as much as possible. ComSR has two trade-off algorithms, ComSR$_\text{non}$ and ComSR$_\text{ful}$, based on whether the database can be completely compressed. Experiments done on a real dataset with different thresholds show that a set of compact and meaningful sequential rules can be found. This shows that the proposed method works.
\end{abstract}


\IEEEpeerreviewmaketitle

\section{Introduction}

\IEEEPARstart{N}{owadays}, the development of the Internet is very fast, and various industries are inseparable from data, such as the large number of transactions and stock fluctuation data generated every day in a wide range of applications. How can we extract more valuable information from the massive amounts of data? Data mining technology \cite{gan2017data,gan2019survey,fournier2022pattern} is a useful tool and has been widely used for big data processing and analysis. The Internet generates massive sequential data (e.g., access logs, purchase behavior, and tour tracks) which can help design makers finish the decision. However, it is hard to balance information redundancy and information security in data technologies. In the meantime, it is not easy to deal with different types of characteristic data in a unified manner. For example, ``excessive data'', ``information explosion'', and ``lack of knowledge'' require us to master more advanced technologies to store and operate on massive data. Data mining \cite{lakshmi2011conceptual} combines statistics, artificial intelligence, databases, and other technologies \cite{klass2019data}. The functions of data mining include but are not limited to feature description, association analysis, data classification, prediction, clustering, and bias analysis. Association analysis studies the co-occurrence frequency of items in the database and identifies frequent itemsets based on two fundamental metrics: \textit{support} and \textit{confidence}. The \textit{confidence} means that a conditional probability of an item appearing in a transaction is used to pinpoint association rules when another item appears.

Sequential rule mining (SRM) \cite{kour2017sequential,huang2022us} is a common data mining task. It introduces the concept of sequential sequence on the basis of association rules. That is, its rules also have sequential relationships. SRM has also been applied in different fields. Researchers can further obtain valuable information through sequential relationships. Setiawan \textit{et al.} \cite{setiawan2018improved} used sequential rules with time constraints to determine the potential local behaviors. Through these behaviors, it will be easier for stakeholders to determine the related behaviors of human capital and productivity improvement. Hus{\'a}k \textit{et al.} \cite{husak2020predictive} used the sequential rules mined by the SRM algorithm to predict security events, which they utilized to create a predictive blacklist. In order to support destination-marketing organizations (DMOs) to promote suitable destinations to potential travelers, the study \cite{vu2018travel} analyzes the behavior of travelers through geotagged photos and one SRM algorithm. In a given sequence database, the user sets \textit{minsup} (the minimum support) and \textit{minconf} (the minimum confidence) thresholds. Then, the SRM algorithm is capable of mining items in each sequence that meet two thresholds at the same time. The discovered rules indicate that if the itemset $X$ appears, the itemset $Y$ will appear with a given confidence, and they can then form a sequential rule $X \to Y$. However, in many cases, users usually have trouble when setting \textit{minsup} and \textit{minconf} thresholds. Compressing interesting patterns (e.g., itemsets \cite{xin2005mining,vreeken2011krimp}, sequential patterns \cite{lam2014mining}) has been studied and is still a challenging task \cite{chandola2007summarization,galbrun2022minimum}. Whether the thresholds are too high or too low, it can result in an information deficiency or an explosion. For example, in a large-scale sequence database, when the user-given thresholds are lower, the number of discovered sequential rules may increase rapidly. It is difficult to extract representative rules from a large set of results. On the other hand, if the two thresholds are larger, the number of the discovered sequential rules is small, and users are likely to be unable to obtain the right information. Therefore, if a certain sequential rule set has a better description of the sequence database than the collection of all frequent sequential rules, then we can find all sequential rules that produce a good description of the sequence database and form a good description of the sequence database. In order to determine whether a sequential rule produces good compression for a sequence database, the Minimum Description Length (MDL) principle \cite{grunwald2005minimum} works very well. In other words, the MDL principle provides a fair method to balance the size of the compressed database and the size of the coding scheme. The so-called minimum description length is related to the final size of the data, which consists of the size of the compressed data and the compressed data model. The MDL principle has the best compression effect.

To address the above limitations, this paper proposes an algorithm, named \textbf{Com}pressing \textbf{S}equential \textbf{R}ule miner (simplified as ComSR), for identifying representative rules from sequence databases. Besides, a sequence database with chronological order in which only a single event can appear at a given time is introduced. To effectively discover sequential rules, we combine frequent sequential rules with the MDL principle. The ComSR algorithm aims to find a set of compact and high-quality sequential rules for extracting compressed and meaningful information from massive data. The major contributions of this paper are concluded as follows:

\begin{itemize}
    \item Based on the MDL principle, we design the coding scheme to compress the sequential rule set and propose the ComSR$_\text{non}$ algorithm to find a set of compact sequential rules from multiple sequences. To our knowledge, this is the first to encode an entire database via sequential rules, while the COSSU algorithm \cite{bourrand2021discovering} finds sequential rules from a long sequence.
    
    \item To address the problem of low compression ratio in ComSR$_\text{non}$, we further propose the ComSR$_\text{ful}$ algorithm by using all 1 $\times$ 1-sized sequential rules with \textit{support} greater than 0 as the basic code set.
    
    \item We perform several experiments for ComSR$_\text{ful}$ and ComSR$_\text{non}$ on a real-life dataset. The final results indicate that both algorithms can compress the set of sequential rules. In particular, ComSR$_\text{ful}$ incurs more execution time than ComSR$_\text{non}$, and ComSR$_\text{non}$ results in lower compression.
\end{itemize}

The rest of this paper is organized as follows. Related works are discussed in Section \ref{sec:relatedwork}. The preliminaries are introduced in Section \ref{sec:preliminaries}. The proposed coding scheme and algorithms are described in Section \ref{sec:CSaA}. The experimental evaluation is performed in Section \ref{sec:experiments}. Finally, we summarize the conclusion and discuss our future work in Section \ref{sec:conclusion}.
\section{Related Work}  \label{sec:relatedwork}

The research presented in this paper is based on sequential rule mining and the MDL principle. A heuristic algorithm is designed to obtain a set of compact and meaningful sequential rules as much as possible. Therefore, we review some advances in the two research fields.

\subsection{Sequential Rule Mining}

Note that the concept of sequential pattern \cite{gan2019survey} is different from that of sequential rule. Zaki \cite{zaki2001spade} proposed the RuleGen algorithm, which utilizes the naive method (i.e., enumeration) to generate sequential rules from two different sequential patterns. What's more, the longer the traversed sequential patterns are, the higher the number of the sequential rules will be obtained. Later, Fournier-Viger \textit{et al.} \cite{fournier2012cmrules} re-named an algorithm as CMDeo and proposed the CMRules algorithm. CMDeo mines sequential rules in a single sequence of events. It is easy to discover common sequential rules from multiple sequences. CMRules mines a more general form of rules such that items in the antecedent and consequent of each rule are unordered. The experimental results show that CMRules is faster and has better scalability than CMDeo under different low \textit{minsup} thresholds.

RuleGrowth \cite{fournier2011rulegrowth} is another algorithm for discovering sequential rules from multiple sequences. Unlike previous algorithms, RuleGrowth utilizes a pattern growth method to discover sequential rules, and it is more efficient and scalable. Then, Fournier-Viger \textit{et al.} \cite{fournier2014erminer} proposed an equivalence class method, called ERMiner, which generates longer sequential rules by continuously merging equivalence classes with the same antecedents or consequences. In addition, the sparse count matrix (SCM) is introduced to prune the search space. In fact, the memory consumption of ERMiner is higher than that of RuleGrowth, which is a space-for-time algorithm. Later, the TRuleGrowth algorithm \cite{fournier2012mining} utilizes window size constraints for mining sequential rules in a sequence database. The SRM problem is also extended to mine the top-$k$ sequential rules in a sequence database. The study \cite{fournier2011mining} proposed a novel method for mining top-$k$ order rules. To reduce redundant results, the TNS algorithm \cite{fournier2013tns} was developed for mining top-$k$ non-redundant sequential rules. Zida \textit{et al.} \cite{zida2015efficient} proposed an algorithm to discover high-utility sequential rules, called HUSRM, and its optimizations reduce execution time and memory usage. Until now, there have only been two algorithms proposed for HUSRM, which are not efficient enough. Recently, Huang \textit{et al.} \cite{huang2022us} proposed a faster algorithm, called US-Rule, which utilizes the rule estimated utility co-occurrence pruning strategy (REUCP) to avoid meaningless computation. In order to reduce the size of the database and the number of scans of the database while using a pattern-growth method, Dalmas \textit{et al.} \cite{dalmas2017twincle} proposed the Time-WINdow, Cost, and LEngth constrained SRM algorithm named TWINCLE, based on the spirit of PrefixSpan \cite{pei2004mining} and HUSRM \cite{zida2015efficient}. Gan \textit{et al.} \cite{gan2022towards} formulated the problem of targeted sequential rule mining and proposed the TaSRM algorithm to obtain all qualified target sequential rules that contain the query items/events. TaSRM can achieve more targeted mining, thereby reducing unnecessary resource consumption.

\subsection{Minimum Description Length}

In the field of data mining, pattern mining is a fundamental task, which mainly includes association rule analysis \cite{gan2017data}, sequential pattern mining \cite{fournier2017survey}, periodic pattern mining \cite{fournier2019efficient}, graph mining \cite{shaul2021cgspan}, high-utility pattern mining \cite{gan2019huopm,gan2020survey}, sequential rule mining \cite{fournier2014erminer}, and so on. There are also many methods for data compression and data mining technologies, and most of them belong to the pattern mining domain \cite{galbrun2022minimum}. Usually, we want to extract a concise and precise set of high-quality results (e.g., itemsets, rules, graphs).

Heierman \textit{et al.} \cite{heierman2003improving} proposed a framework based on the Episode Discovery (ED) algorithm combined with the MDL principle. It discovers behavioral patterns within a sequential data stream. Keogh \textit{et al.} \cite{keogh2004towards,keogh2007compression} proposed a compression-based dissimilarity measure (CDM), which evaluates the relative gain in compressing two strings that are concatenated rather than separated. Inspired by Kolmogorov complexity, this metric is used to combat the large number of parameters when mining sequential rules and can be adopted in any off-the-shelf compression algorithm. Faloutsos \textit{et al.} \cite{faloutsos2007data} argued that the close relation between data mining, compression, and Kolmogorov complexity by a unified theory of data mining had little hope. Siebes \textit{et al.} \cite{siebes2006item} combined the MDL principle to solve the combinatorial explosion problem; that is, the most frequent itemset is the one that best compresses the database. At the same time, Siebes \textit{et al.} \cite{leeuwen2006compression} introduced another algorithm based on the MDL principle to mine small but high-quality itemsets. Their achievements led to a series of fruitful research, including adaptation to different assignments, algorithm improvement, and all kinds of applications of the original algorithm \cite{vreeken2011krimp}. Then, Smets \textit{et al.} \cite{smets2012slim} modified the Krimp algorithm \cite{vreeken2011krimp} and proposed the Slim algorithm, which generates candidates greedily by merging patterns but does not evaluate candidates from a pre-mined list. Hess \textit{et al.} \cite{hess2014shrimp} proposed the Shrimp algorithm to construct an FP-tree-like data structure that reflects the impact of schema selection on the database and the quality of the selection can be calculated more quickly. Thus, this Shrimp algorithm can more easily recalculate usages when the coding scheme is updated. Sampson and Berthold \cite{sampson2014widened} introduced Reverse Standard Candidate Order (RSCO) as Krimp's candidate ranking heuristic. Lam \textit{et al.} \cite{lam2014mining} proposed a heuristic algorithm which is based on the MDL principle to mine compressed patterns.

Previous studies \cite{simovici2013minability,simovici2015compression} showed that compression can be used to evaluate interesting mining processes and similarity between complex objects. They focused on studying the compressibility of symbolic strings and the usage of compression when computing similarity in text corpora. The term ``compression'' used by Chandola \textit{et al.} \cite{chandola2007summarization} is a metaphor rather than a practical tool. Their proposed method for mining patterns from categorical attribute tables relies on a pair of ad hoc scores, which intuitively measure how well the considered set of patterns compresses the data and how much information is lost. COSSU \cite{bourrand2021discovering} continuously constructs a sequential rule set as a code set describing the length through a greedy strategy algorithm.
\section{Preliminaries and Problem Definition}
\label{sec:preliminaries}

\subsection{Basic Concepts}
\label{subsec:concepts}

Let $I$ = \{$i_1$, $i_2$, $i_3$, $\dots$, $i_l$\} be a set of items, where an itemset is a set of unordered and non-repeating items, expressed as $I_x$ = \{$i_1$, $i_2$, $i_3$, $\dots$, $i_m$\} $\subseteq$ $I$. Note that the elements $i_m$ in the itemset $I_x$ do not have a strict order. A \textit{sequence s} is an ordered list of itemsets $s$ = $<$$I_1$, $I_2$, $\dots$, $I_n$$>$ where $I_k$ $\subseteq$ $I$ (1 $\leq$ $k$ $\leq$ $n$). A sequence database is a set of sequences $\mathcal{D}$ = $<$\textit{seq}$_1$, \textit{seq}$_2$, $\dots$, \textit{seq}$_p$$>$, where each sequence (\textit{seq}) has an identification number (SID). As shown in Table \ref{tab:seq_db}, a sequence database contains 4 sequences. Each letter represents an item, and the curly brackets represent an itemset. For example, the \textit{seq}$_2$ contains 4 itemsets, and the first itemset means that both $a$ and $d$ appear at the same time, and the second itemset means that it appears next $\{a, d\}$. In this paper, we use a special sequence database. That is, each itemset only contains one item. In other words, at each time point of a sequence (e.g., \textit{seq}$_3$), only one item or one transaction occurs.

\begin{table}[h]
	\centering
	\caption{Sequence database}
	\label{tab:seq_db}
	\begin{tabular}{|c|c|}  
		\hline 
		\textbf{SID} & \textbf{Sequence} \\
		\hline  
		\(seq_{1}\) & $<$\{\textit{a}, \textit{g}\}, \{\textit{c}\}, \{\textit{g}\}, \{\textit{e}, \textit{b}\}$>$ \\ 
		\hline
		\(seq_{2}\) & $<$\{\textit{a}, \textit{d}\}, \{\textit{c}\}, \{\textit{b}\}, \{\textit{g}, \textit{b}, \textit{e}, \textit{f}\}$>$ \\ 
		\hline
		\(seq_{3}\) & $<$\{\textit{f}\}, \{\textit{a}\}, \{\textit{g}\}, \{\textit{b}\}$>$ \\ 
		\hline
		\(seq_{4}\) & $<$\{\textit{a}, \textit{g}\}, \{\textit{f}, \textit{g}, \textit{h}\}$>$ \\ 
		\hline
	\end{tabular}
\end{table}

The sequential rule is the connection between two unordered itemsets, which is represented by $I_x$ $\to$ $I_y$ ($I_x \cap I_y$ = $\varnothing$ and $I_x$, $I_y$ $\neq$ $\varnothing$). $I_x$ represents the antecedent of $I_x$ $\to$ $I_y$, and $I_y$ represents the consequent. Besides, the itemsets in the antecedent and consequent are both unordered. 

\begin{table}[h]
	\centering
	\caption{Some sequential rules in Table \ref{tab:seq_db}}
	\label{tab:SR_in_tab1}
	\begin{tabular}{|c|c|c|c|}  
		\hline 
		\textbf{ID} &\textbf{Rule} & \textbf{Support} & \textbf{Confidence}\\
		\hline  
		\(rule_{1}\) & \{$\textit{a}$\}$\to$\{$\textit{c}$\} & 0.5 &0.5\\ 
		\hline
		\(rule_{2}\) & \{$\textit{a}$, $\textit{c}$\}$\to$\{$\textit{g}$, $\textit{b}$\} & 0.5 & 1.0\\ 
		\hline
		\(rule_{3}\) & \{$\textit{a}$\}$\to$\{$\textit{b}$\} & 0.75 & 0.75\\ 
		\hline  
        \(rule_{4}\) & \{$\textit{a}$, $\textit{g}$\}$\to$\{$\textit{b}$\} & 0.5 & 0.5\\ 
		\hline
		\(rule_{5}\) & \{$\textit{a}$, $\textit{f}$\}$\to$\{$\textit{g}$\} & 0.25 & 0.33\\ 
		\hline
	\end{tabular}
\end{table}

Considering a sequential rule $I_x$ $\to$ $I_y$, its size is $p \times q$ iff $|I_x|$ = $p$ and $|I_y|$ = $q$. Therefore, the smallest sequential rule is 1 $\times$ 1 size. If the size of another sequential rule is $m \times n$, then the former is larger than the latter if only $p$ $>$ $m$ and $q \geq n$ or $p \geq m$ and $q$ $>$ $n$ must be satisfied. In Table \ref{tab:SR_in_tab1}, the size of \textit{rule}$_2$ is $2 \times 2$. The \textit{support} of a rule is a vital metric to evaluate the interest of the rule. The definition is listed as follows:

\begin{equation}
	\textit{sup}_{\mathcal{D}}(\textit{rule}) = \frac{\left|{seq|seq \in \mathcal{D} \land rule \subseteq seq}\right|}{\left| \mathcal{D} \right|}.
\end{equation}

\textit{seq}$_1$ and \textit{seq}$_2$ are two sequences that satisfy the \textit{rule}$_1$, and a sequence database contains four sequences. Then, \textit{sup}$_{\mathcal{D}}$(\textit{rule}$_1$) is 2/4 = 0.5. In addition, the \textit{confidence} of a rule indicates the probability of the rule appearing when the antecedent of the rule appears. The calculation formula is shown as follows:

\begin{equation}
    \textit{conf}_{\mathcal{D}}(\textit{rule}) = \frac{ \left|{seq|seq \in \mathcal{D} \land rule \subseteq seq}\right|}{ \left|{seq|seq \in \mathcal{D} \land I_x \subseteq seq}\right|}.
\end{equation}

As an example, consider the \textit{rule}$_1$. The \textit{rule}$_1$ $\{a\}$ $\to$ $\{c\}$ occurs in \textit{seq}$_1$ and \textit{seq}$_2$. The antecedent $\{a\}$ of \textit{rule}$_1$ is also satisfied by four sequences (\textit{seq}$_1$, \textit{seq}$_2$, \textit{seq}$_3$ and \textit{seq}$_4$). Then, \textit{conf}$_{\mathcal{D}}$(\textit{rule}$_1$) is 2/4 = 0.5.

\begin{figure*}[h]
	\centering
	\includegraphics[scale=0.98]{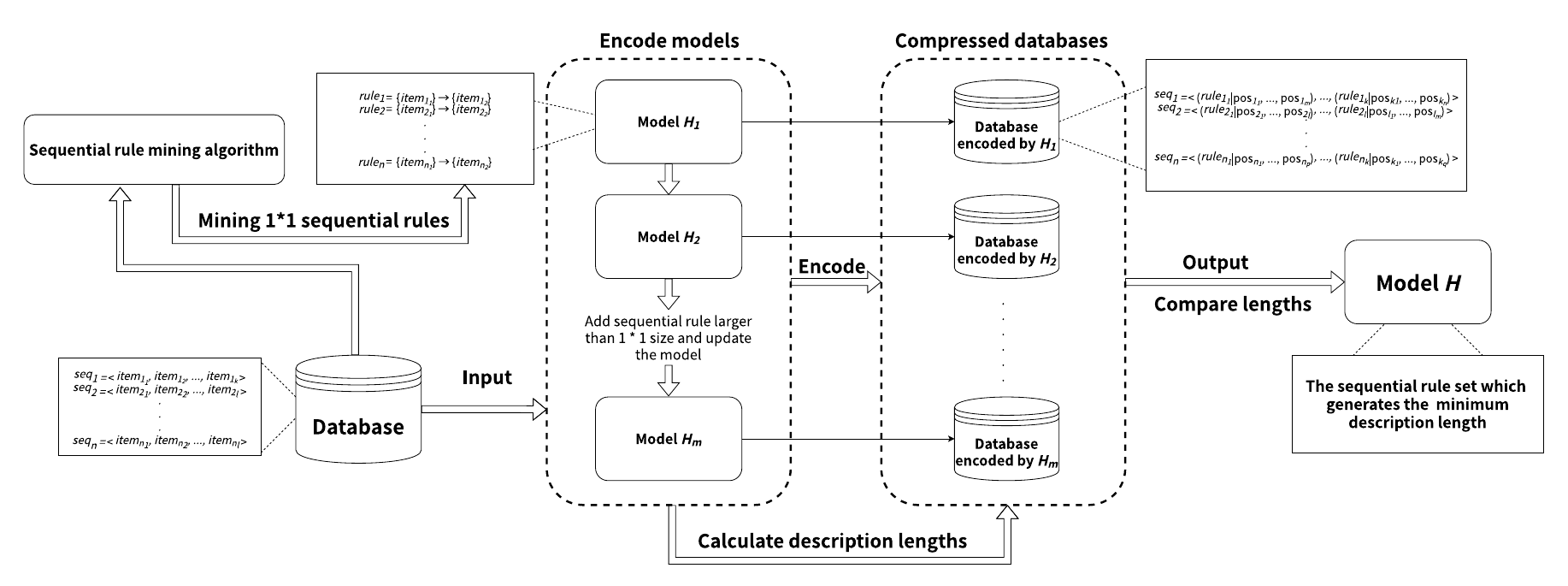}
	\caption{Framework of the ComSR algorithm.}
	\label{fig:framework}
\end{figure*}

\subsection{Minimum Description Length Principle}

The Minimum Description Length (MDL) principle is an information theory-based standard \cite{rissanen1978modeling}. The MDL's basic principle is that for a given model set $\mathcal{H}$ and dataset $\mathcal{D}$, select the model $H \in \mathcal{H}$ that can produce the best compression for $\mathcal{D}$. In order to save storage space, the model $H \in \mathcal{H}$ is used to encode and compress $\mathcal{D}$. $H$ and $\mathcal{D}$ will be saved concurrently in order to restore the data later. Therefore, the total saved data length \textit{Length} is equal to the length $L(\mathcal{D}|H)$ of the dataset after encoding and compression plus the data length $L(H)$ required to save the model. The MDL principle satisfies the following formula:

\begin{equation}
    \label{formula:length}
	Length = \arg\min_{H \in \mathcal H} {L(\mathcal{D}|H)+L(H)}.
\end{equation}

Here, a model $H$ that minimizes the description length is calculated in all model sets $\mathcal{H}$. The model encodes and compresses the data, and then the final result is the best. In this paper, a sequence database is used, and $L(\mathcal{D}|H)$ in formula \ref{formula:length} refers to the sequence database composed of the collection of each sequence \textit{seq} in $\mathcal{D}$. The sequence database of each sequence is determined by the model $H$. The length of the representative is $L(seq|H)$. After encoding all the sequences in the sequence database using the model $H$, $\mathcal{D}|H$ is obtained, and its length is $L(\mathcal{D}|H)$. That is, $L(\mathcal{D}|H) $ = $\sum_{seq \in \mathcal{D}}{L(seq|H)}$. Then the encoding formula for the sequence database can be defined:

\begin{equation}
	\textit{Length} = \arg\min_{H \in \mathcal H} {\sum_{seq \in \mathcal{D}}{L(seq|H)+L(H)}}.
\end{equation}

In Section \ref{subsec:coding_and_alg_bg}, a more detailed introduction will be carried out according to the specific situation of the coding scheme. Herein, we describe the framework of our study (Fig. \ref{fig:framework}). Firstly, we will use a special sequence database as input. The size of an itemset of a sequence in the sequence database is 1. We also use a sequential rule mining algorithm to extract a set of sequential rules (1 $\times$ 1) from the sequence database as the initial encoding model. Then, the sequential rules of other sizes will be taken as candidate rules to continuously update the encode model. Section \ref{subsec:coding_and_alg_bg} describes the coding scheme. The heuristic algorithms introduced in Section \ref{subsec:proposed_alg} and Section \ref{subsec:improve_compression_ratio} will compare the description length of different models on the compressed database and find the unique encode model which makes the database with the minimum description length as much as possible as output. The found encode model is a set of compact and meaningful sequential rules.

\section{Coding Scheme and Algorithms}
\label{sec:CSaA}

\subsection{Coding and Algorithm Background}
\label{subsec:coding_and_alg_bg}

The study \cite{lam2014mining} uses pointers to represent the coding of sequential patterns (i.e., a dictionary-like coding table). If there are items in a sequence that meet the conditions of the encoding rules, the items will be replaced with pointers. If the remaining items in the same sequence cannot be encoded, they will be calculated according to the original length. The following Table \ref{tab:db_dic} \cite{lam2014mining} is a sequence database and pointer dictionary exemplified in \cite{lam2014mining}. For example, \{$c$, $a$, $b$\} in $seq_2$ can be represented by the pointer 2 in Table \ref{tab:db_dic}, and $seq_2$ only needs two pointers to complete the encoding. Inspired by the study \cite{lam2014mining}, we represent the items in the sequence by sequential rule pointers plus corresponding indexes or timestamps. The coding scheme will combine the sequential rules and the indexes or timestamps of items in the sequence to encode the sequence. In this way, the compressed data can be accurately recovered.

\begin{table}[h]
	\centering
	\caption{A database and a pointer dictionary}
	\label{tab:db_dic}
	\begin{tabular}{|c|c|}  
		\hline 
		\textbf{Database $\mathcal{D}$} & \textbf{Dictionary \textit{H}}  \\
		\hline  
		$seq_{1}$ = (\textit{a},1)(\textit{c},3)(\textit{a},4)(\textit{b},7)(\textit{c},8)(\textit{b},9) & pointer 1: \textit{a}\textit{b} \\ 
		\hline
		$seq_{2}$ = (\textit{c},2)(\textit{a},4)(\textit{b},5)(\textit{c},6)(\textit{a},7)(\textit{b},8) & pointer 2: \textit{c}\textit{a}\textit{b} \\  
		\hline  
	\end{tabular}
\end{table}

The research \cite{siebes2006item} adopts a heuristic method to seek an itemset as much as possible, which is different from the sequential rule set in this paper. At the same time, the itemset mined by the algorithm \cite{siebes2006item} is not limited by the sequential relationship. It is not necessary to consider the sequential relationship between the patterns in the search space, and thus the processing is relatively simple. The algorithm \cite{siebes2006item} first uses all the items in the database as the initial code set, and then takes the remaining frequent itemsets as the candidate code set. A \textit{cover} index is used to sort the candidate code set, so that itemset of the larger \textit{cover} will be encoded. Then the algorithm \cite{siebes2006item} calculates the description length, and it compares the description length after adding the candidate pattern with the description length not added. If the description length becomes large, the candidate pattern is added to the basic code set; otherwise, the next round of traversal is performed. This operation is continued until the candidate pattern set is empty. Based on the heuristic algorithm and the MDL principle, this paper proposes a coding scheme to encode the sequence database and discovers a set of compact and meaningful sequential rules.

\subsection{Coding Scheme}
\label{subsec:code_scheme}

In this section, we will show how to use sequential rules to encode sequence databases. Data compression saves storage space by finding regularities in data and reducing data size. These regularities usually correspond to significant patterns in the data. For example, the regularity that one thing happens after another may be repeated many times, which is the same as the representation of sequential rules. Thus, this regularity can be used to compress data, and can be discovered by SRM algorithms. However, there are several differences between the SRM algorithm and data compression. Data compression looks for any type of regularity, and its goal is to reduce the size of the data representation. From the point of view of SRM, not all frequent sequential rules can describe the data completely.

The discovered rules of the ComSR algorithm are used to indicate a sequential relationship between the occurrence of two frequent disorder items in time. That is, if an itemset $X$ appears before an itemset $Y$, it means $X \to Y$ (the items in $X$ and $Y$ are disordered). Therefore, after the database is encoded by different sequential rules, this paper only considers the comparison of different description lengths of the database in the calculation process and thus achieves the final goal through this principle. The coding scheme is introduced below. For a given sequence database $\mathcal{D}$, we can use different sizes of sequential rules as the code set (i.e., the model $H$). The adopted dictionary mode to encode and decode is proposed in Ref. \cite{galbrun2018mining}. Tables \ref{tab:data_codeset_h1} and \ref{tab:data_codeset_h2} take different sequential rule sets as the code sets to encode the same sequence database.

\begin{table}[h]
	\centering
	\caption{The uncoded database and code set $H_{1}$}
	\label{tab:data_codeset_h1}
	\begin{tabular}{|c|c|}  	
		\hline 
		\textbf{Database $\mathcal{D}$} & \textbf{Dictionary $H_{1}$} \\
		\hline  
		 \textit{seq}$_1$ = $<$\{\textit{a}\},\{\textit{b}\},\{\textit{c}\},\{\textit{d}\},\{\textit{e}\},\{\textit{f}\}$>$  & \textit{rule}$_1$ = \{\textit{a}\}$\to$\{\textit{b}\} \\ 
		\hline
		\textit{seq}$_2$ = $<$\{\textit{a}\},\{\textit{b}\},\{\textit{d}\},\{\textit{e}\},\{\textit{g}\},\{\textit{f}\}$>$ & \textit{rule}$_2$ = \{\textit{d}\}$\to$\{\textit{e}\}\\ 
		\hline
		\( - \) & \textit{rule}$_3$ = \{\textit{c}\}$\to$\{\textit{f}\}\\ 
		\hline
		\( - \) & \textit{rule}$_4$ = \{\textit{g}\}$\to$\{\textit{f}\}\\ 
		\hline
	\end{tabular}
\end{table}

\begin{table}[h]
	\centering
	\caption{The uncoded database and code set $H_{2}$}
	\label{tab:data_codeset_h2}
	\begin{tabular}{|c|c|}  
		\hline 
		\textbf{Database $\mathcal{D}$} & \textbf{Dictionary $H_{2}$} \\
		\hline  
		\textit{seq}$_{1}$ = $<$\{\textit{a}\},\{\textit{b}\},\{\textit{c}\},\{\textit{d}\},\{\textit{e}\},\{\textit{f}\}$>$  & \textit{rule}$_{1}$ = \{\textit{a},\textit{b}\}$\to$\{\textit{d},\textit{e}\} \\ 
		\hline
		\textit{seq}$_{2}$ = $<$\{\textit{a}\},\{\textit{b}\},\{\textit{d}\},\{\textit{e}\},\{\textit{g}\},\{\textit{f}\}$>$  & \textit{rule}$_{2}$ = \{\textit{c}\}$\to$\{\textit{f}\}\\ 
		\hline
		\( - \) & \textit{rule}$_{3}$ = \{\textit{g}\}$\to$\{\textit{f}\}\\ 
		\hline
	\end{tabular}
\end{table}

Now we use the code sets $H_1$ and $H_2$ to represent each sequence in the database (i.e., Table \ref{tab:data_codeset_h1} and Table \ref{tab:data_codeset_h2}). First, it should be determined whether a rule conforms to the order of occurrence of items in the sequence. If so, this rule will be used to represent all these items because the antecedents and consequences of the rules are unordered. Then, subscripts are added to ensure the smooth subsequent decoding process of the unordered itemsets. As shown in Tables \ref{tab:encode_data_h1} and \ref{tab:encode_data_h2}, as long as all items of the sequential rules in the code set include all items of the database, the database can be represented fully by a set of sequential rules. Using model $H_1$ as an example, if \{$a$\} in \textit{seq}$_{1}$ appears before \{$b$\}, then \{$a$\} and \{$b$\} can be represented by \textit{rule}$_1$ in model $H_1$, and (1, 2) are two-position subscripts. If the sequence database also has time points, as shown in Table \ref{tab:db_dic}, the time points are added to indicate the position. The remaining items \{$c$\}, \{$d$\}, \{$e$\}, \{$f$\} can be represented by $\textit{rule}_2$ and $\textit{rule}_4$, respectively. The subscripts corresponding to the two sequential rules are (4, 5) and (3, 6). Then \textit{seq}$_{1}$ can be expressed as $<$($\textit{rule}_1$|1, 2), ($\textit{rule}_2$|4, 5), ($\textit{rule}_3$|3, 6)$>$ after encoding. $seq_2$ is encoded by model $H_2$ and expressed as $<$($\textit{rule}_1$|1, 2, 3, 4), $(\textit{rule}_3$|5, 6)$>$ using the same manner.

\begin{table}[h]
	\centering
	\caption{The database encoded by the code set $H_1$}
	\label{tab:encode_data_h1}
	\begin{tabular}{|c|c|}
		\hline 
	    & \textbf{Database $\mathcal{D}|H_1$} \\
		\hline  
		\textit{seq}$_{1}$ & $<$(\textit{rule}$_{1}$|1,2), (\textit{rule}$_{2}$|4,5), (\textit{rule}$_{3}$|3,6)$>$ \\ 
		\hline
		\textit{seq}$_{2}$ & $<$(\textit{rule}$_{1}$|1,2), (\textit{rule}$_{2}$|3,4), (\textit{rule}$_{4}$|5,6)$>$)\\ 
		\hline						
	\end{tabular}
\end{table}

\begin{table}[h]
	\centering
	\caption{The database encoded by the code set $H_2$}
	\label{tab:encode_data_h2}
	\begin{tabular}{|c|c|}  	
		\hline 
		& \textbf{Database $ \mathcal{D}|H_2$} \\
		\hline  
		\textit{seq}$_{1}$   & $<$(\textit{rule}$_{1}$|1,2,4,5), (\textit{rule}$_{2}$|3,6)$>$ \\ 
		\hline
		\textit{seq}$_{2}$   & $<$(\textit{rule}$_{1}$|1,2,3,4), (\textit{rule}$_{3}$|5,6)$>$\\ 
		\hline						
	\end{tabular}
\end{table}

Next, we analyze the difference in description length between different sequential rule sets as code sets after coding the sequence database. We set the unit of description length as unit 1, and calculate the description length after encoding the database with different sequential rule sets. If the final description length is reduced after merging the small-sized sequential rules into the large-sized sequential rules, the sequence database will have a smaller description length represented by the large sequential rules with a certain degree of \textit{support} and \textit{confidence}.

In addition, it is necessary to understand the length of the encoded model, which includes the number of sequential rules plus the total number of items in all sequential rules. For example, the length of the model $H_1$ is $L(H_1)$ = $\left| H \right|$ + $\sum_{R\in \mathcal{R}}{\left| R \right|}$ = 4 + (2 + 2 + 2 + 2) = 12, where $\left| H \right|$ represents the number of sequential rules. Each rule is represented by a quantity unit. $\left| R \right|$ is the number of items in the antecedent of the rule $R$ plus the number of items in the consequent of the rule $R$. Thus, the length of $H_2$ is $L(H_2)$ = 3 + (4 + 2 + 2) = 11. The length of the database encoded with $H_1$ is $L(\mathcal{D}|H_1)$ = $\sum{f_{rule}\times (\left| R \right| + 1)}$ = 2 $\times$ (2 + 1) + 2 $\times$ (2 + 1) + 1 $\times$ (2 + 1) + 1 $\times$ (2 + 1) = 18. This length means that $rule_1$ is encoded twice and needs 2 quantity units to express its position and 1 quantity unit to express the rule. Similarly, the length of the database encoded with $H_2$ is $L(\mathcal{D}|H_2)$ = 2 $\times$ (4 + 1) + 1 $\times$ (2 + 1) + 1 $\times$ (2 + 1) = 16. Finally, the description length of the sequence database encoded by the $H_1$ model is larger than that encoded by the $H_2$ model, e.g., $L(H_1)$ + $L(\mathcal{D}|H_1)$ = 12 + 18 = 30 $ >$ $L(H_2)$ + $L(\mathcal{D}|H_2)$ = 11 + 16 = 27. In this paper, this coding method is used in combination with the algorithm, as presented in the next section. It can discover a set of more compact and meaningful sequential rules.

\subsection{Proposed Algorithm}
\label{subsec:proposed_alg}

When the user-given \textit{minsup} and \textit{minconf} are very small, the scale of the resulting sequential rule set will become very large. It is impractical to use a huge sequential rule set as the code set. In addition, it is also very difficult to select the best sequential rule set as the code set to encode the database. Therefore, this paper utilizes an effective heuristic algorithm to solve this issue. 

\textbf{Find a set of sequential rules}. Algorithm \ref{alg:ComSRnon} describes the ComSR$_\text{non}$ (MDL-based Compressing Sequential Rules algorithm that does not fully compress the database) algorithm. It takes a sequence database (this sequence database is a special case, e.g., Table \ref{tab:data_codeset_h1}), the minimum \textit{support} and minimum \textit{confidence} as input, and a set of sequential rules $R$ as output. ComSR$_\text{non}$ first uses one of the existing SRM algorithms to mine a set of sequential rules $\mathbb R$ with the specified thresholds (Line 1). The procedure initializes the basic code set (\textit{CodeSet}), and then extracts the 1 $\times$ 1-sized sequential rules from the sequential rule set $\mathbb R$ as the basic code set \textit{CodeSet} (Line 2). The sequential rules greater than 1 $\times$ 1 in $\mathbb R$ are selected as the set of candidate rules (\textit{CandidateRules}), that is sorted in descending order of \textit{support} size (Line 3). Then, the candidate code set (\textit{canCodeSet}), copies the current code set \textit{CodeSet} and adds a new candidate sequential rule (\textit{candidate}) from the candidate rule set \textit{CandidateRules} (Lines 4--7). The candidate code set \textit{canCodeSet} is first sorted in descending order of rule size and then sorted in descending order of \textit{support} size (Line 8). The algorithm compares the description length of the code set \textit{CodeSet} and the candidate code set \textit{canCodeSet} after encoding the database. If the description length becomes smaller, update the code set \textit{CodeSet} (Lines 9 and 10). The same operation is performed every time until the candidate rule set \textit{CandidateRules} is empty. After the last loop, the ComSR$_\text{non}$ algorithm returns the final code set (Line 11).

\begin{algorithm}[h]
	\caption{ComSR$_\text{non}$ algorithm}
	\label{alg:ComSRnon}
    \LinesNumbered
	\KwIn{$\mathcal{D}$: a sequence database; \textit{minsup}: the minimum \textit{support}; \textit{minconf}: the minimum \textit{confidence}.} 
	\KwOut{\textit{R}: a set of compact and meaningful sequential rules \textit{R}.}
	
	$\mathbb R \leftarrow$  \textit{MineSequenceRule}($\mathcal{D}$, \textit{minsup}, \textit{minconf});
	
	\textit{CodeSet} $\leftarrow$ \textit{InitialCode}($\mathbb R$);
	
	\textit{CandidateRules} $\leftarrow$ \textit{sort\_descending}($\mathbb R$ - \textit{CodeSet}, \textit{support}(\textit{rule}));
	
	\While{\textit{CandidateCodeSet} $\neq \varnothing$}{
		\textit{candidate} $\leftarrow$ \textit{CandidateRules}.\textit{pop}(0);
		
		\textit{canCodeSet} $\leftarrow$ \textit{CodeSet};
		
		\textit{canCodeSet} $\leftarrow$ \textit{canCodeSet} $\cup$ \textit{candidate};
		
		\textit{sort\_descending}(\textit{canCodeSet}, $[$\textit{size}(\textit{rule}), \textit{support}(\textit{rule})$]$);
		
		\If{\textit{CompressLength}(\textit{canCodeSet}, $\mathcal{D}$) $<$ \textit{CompressLength}(\textit{CodeSet}, $\mathcal{D}$)}{
				\textit{CodeSet} $\leftarrow$ \textit{canCodeSet};
		}
	}
	
	\textbf{return} \textit{CodeSet}	
\end{algorithm}

\begin{algorithm}[h]
	\small
	\caption{CompressLength algorithm}
	\label{CL:CL}
	\LinesNumbered
	\KwIn{$\mathcal{D}$: a sequence database; \textit{CodeSet}.} 
	\KwOut{$L(H)+L(\mathcal{D}|H)$: a compact and meaningful set of sequential rule set \textit{R}.}		
	\For{\textit{rule} $\in$ \textit{CodeSet}}{
		\For{\textit{seq} $\in$ $\mathcal{D}$}{
		\If{\textit{rule} \rm in \textit{seq}}{
			remove all \textit{rule}.items in \textit{seq};\\
			\textit{freq\_rule} $\leftarrow$ \textit{freq\_rule} + 1; \\
		}
		}
	}
	\For{\textit{seq} $\in$  $\mathcal{D}$}{
	    \For{\textit{rule} $\in$  \textit{CodeSet}}{
	        \If{\textit{seq}.\rm length == 1}{
	            \uIf{\textit{rule}.\rm size == 1$\times$1 and \textit{rule.Antecedent} == \textit{seq} and \textit{rule.Consequence} in \textit{seq}$_\textit{SID}$}{
	            remove \textit{rule.Antecedent} in \textit{seq};\\
	            \textit{freq\_rule} $\leftarrow$ \textit{freq\_rule} + 1;\\
	            }
	            \uElseIf{\textit{rule}.\rm size == 1$\times$1 and \textit{rule.Consequence} == \textit{seq} and \textit{rule.Antecedent} in \textit{seq}$_\textit{SID}$}{
	            remove \textit{rule.Consequence} in \textit{seq};\\
	            \textit{freq\_rule} $\leftarrow$ \textit{freq\_rule} + 1; \\
	            }
	        }
	    }
	}
	\textbf{return} $L(H)+L(\mathcal{D}|H)$	
\end{algorithm}

\textbf{Calculate the description length}. Algorithm \ref{CL:CL} is the process of compressing and encoding the sequence database with the code set and calculating the description length. The input of the algorithm includes $\mathcal{D}$ and \textit{CodeSet}. First, Algorithm \ref{CL:CL} traverses the sequential rules in the code set \textit{CodeSet} (Line 1). Then the procedure traverses each sequence in the database (Line 2) to judge whether the sequential rule exists in the current sequence. If yes, Algorithm \ref{CL:CL} deletes and records the number of occurrences of the sequential rule for subsequent calculations (Lines 4--6). Since a single item may not be encoded during the encoding of the sequence database, we designed an additional loop to traverse the uncoded items in the sequence. If there is only a single item left in the sequence, we will encode it with a 1 $\times$ 1 sequential rule. In addition, the sequential rule used for encoding also needs to exist in the uncoded sequence. First, Algorithm \ref{CL:CL} traverses the uncoded sequence in the database (Line 9). Then, all sequential rules are traversed from the code set \textit{CodeSet} (Line 10). Line 11 judges whether the number of items in the sequence is equal to 1. If so, the procedure continues to judge the next 3 conditions. The first is whether the size of the sequential rule is 1 $\times$ 1, the second is whether the antecedent or consequent of the sequential rule is equal to the item in the sequence, and the third is whether the sequential rule can be mined from the initial sequence (Lines 12--17). If the above three conditions are met, the procedure codes the last item, and then adds 1 to the coding times of the corresponding sequential rule. Finally, the length after calculation is returned through the coding scheme discussed in Section \ref{subsec:code_scheme} (Line 21). In fact, Algorithm \ref{CL:CL} continuously traverses and deletes the items in the qualified sequential rules sorted by rule size and \textit{support} on the sequence database. In this way, the most common sequential rules can be coded to find a set of sequential rules that are both meaningful and small.

\subsection{Improve Compression Ratio}
\label{subsec:improve_compression_ratio}

The initial \textit{CodeSet} may not be able to completely encode the database because of thresholds. The itemsets in the discovered 1 $\times$ 1 sequential rules may not be able to fully cover the itemsets of the database. Therefore, Algorithm \ref{alg:ComSRful} is proposed on this basis to solve this issue. It describes the ComSR$_\text{ful}$ algorithm (MDL-based compressing sequential rules that fully compresses the database). It can fully encode the database. ComSR$_\text{ful}$ focuses on dealing with a special sequence database. If there is a sequence of a single item in the database, the sequence will not be encoded because the antecedents and antecedents of the sequential rules are not empty. Besides, a single item cannot be represented by any sequential rules. The difference between ComSR$_\text{ful}$ and ComSR$_\text{non}$ is shown in Line 2. ComSR$_\text{ful}$ uses all the sequential rules in the sequence database with a \textit{support} greater than 0 and a size of 1 $\times$ 1 as the code set \textit{CodeSet}. The itemset of \textit{CodeSet} can completely cover the itemset of the sequence database. Except for the sequence of the above-mentioned single item in the sequence database, which cannot be encoded, the compression rate can reach a 100\% encoding ratio in most cases.

\begin{algorithm}[h]
	\small
	\caption{ComSR$_\text{ful}$ algorithm}
	\label{alg:ComSRful}
	\LinesNumbered
	\KwIn{$\mathcal{D}$: a sequence database; \textit{minsup}: the minimum \textit{support}; \textit{minconf}: the minimum \textit{confidence}.} 
	\KwOut{\textit{R}: a set of compact and meaningful sequential rules \textit{R}.}		
	$\mathbb R \leftarrow$  \textit{MineSequenceRule}($\mathcal{D}$, \textit{minsup}, \textit{minconf});\\
	\textit{CodeSet} $\leftarrow$ \textit{MineAllOneRule}($\mathbb R$);\\
	\textit{CandidateRules} $\leftarrow$ \textit{sort\_descending}($\mathbb R$ - \textit{CodeSet}, \textit{support}(\textit{rule}));\\
	\While{ \textit{CandidateCodeSet} $\neq \varnothing$}{
		\textit{candidate} $\leftarrow$ \textit{CandidateRules}.\textit{pop}(0);\\
		\textit{canCodeSet} $\leftarrow$ \textit{CodeSet};\\
		\textit{canCodeSet} $\leftarrow$ \textit{canCodeSet} $\cup$ \textit{candidate};\\
		\textit{sort\_descending}(\textit{canCodeSet}, $[$\textit{size}(\textit{rule}), \textit{support}(\textit{rule})$]$);\\
		\If{\textit{CompressLength}(\textit{canCodeSet}, $\mathcal{D}$) $<$ \textit{CompressLength}(\textit{CodeSet}, $\mathcal{D}$)}{
				\textit{CodeSet} $\leftarrow$ \textit{canCodeSet};\\
		}
	}
	\textbf{return} \textit{CodeSet}	
\end{algorithm}

Extracting the 1 $\times$ 1 sequential rules from a sequence database will make the basic code set \textit{CodeSet} larger, which will lead to the final sequential rule set becoming larger. Many sequential rules with low \textit{support} will only be used for one-time coding. Besides, the running time will be greatly increased. In the experimental results, some sequential rules have lower \textit{minsup} and \textit{minconf}. ComSR$_\text{ful}$ can provide a more detailed description of the original sequence database. The improvement of the compression ratio has a certain impact on the final sequential rule set, because the final result may contain sequential rules with \textit{support} or \textit{confidence} lower than the thresholds. However, the final sequential rule set has a more complete description of the original sequence database. The efficiency of the operation will be reduced due to the increase in the code set size, and the runtime consumption will be greatly increased. While improving the compression ratio, it also has other impacts. ComSR$_\text{ful}$ and ComSR$_\text{ful}$ are two distinct trade-offs that can be compared.

In general, algorithms are mainly divided into two types. Both ComSR$_\text{non}$ and ComSR$_\text{ful}$ utilize a heuristic method and the MDL principle to find a set of compact sequential rules. In addition, the CompressLength procedure is capable of calculating the description length of the current sequential rule set that is the code set.
 
\section{Experiments}  \label{sec:experiments}

We conducted experiments to evaluate the performance and scalability in this section. To our knowledge, this is the first algorithm to encode an entire database via sequential rules. The most relevant work is the COSSU algorithm \cite{bourrand2021discovering} which finds sequential rules from a long sequence. Both ComSR$_\text{non}$ and ComSR$_\text{ful}$ are implemented in the Python language with version 3.7.3. The platform is a macOS PC equipped with a 2.6 GHz six-core Intel Core i7 CPU and 16 GB of 2400 MHz DDR4 RAM.

\begin{table}[h]
	\centering
	\caption{Experimental design}
	\label{tab:exp_design}
	\begin{tabular}{|c|c|c|c|}  
		\hline 
		\textbf{Algorithm} & \textbf{\textit{minsup}} & \textbf{\textit{minconf}} & \textbf{Experiments} \\
		\hline  
		ComSR$_\text{non}$ & 0.3 & \textit{variable} & Fig. \ref{fig:ComSR_non_runtime}(a), Fig. \ref{fig:ComSR_non_compression_ratio}(a), Fig. \ref{fig:ComSR_non_SR_size}(a)\\ 
		\hline
		ComSR$_\text{ful}$ & 0.5 & \textit{variable} & Fig. \ref{fig:ComSR_ful_runtime}(a), Fig. \ref{fig:ComSR_ful_SR_size}(a)\\ 
		\hline
		ComSR$_\text{non}$ & \textit{variable} & 0.5 & Fig. \ref{fig:ComSR_non_runtime}(b), Fig. \ref{fig:ComSR_non_compression_ratio}(b), Fig. \ref{fig:ComSR_non_SR_size}(b)\\ 
		\hline
		ComSR$_\text{ful}$ & \textit{variable} & 0.7 & Fig. \ref{fig:ComSR_ful_runtime}(b), Fig. \ref{fig:ComSR_ful_SR_size}(b)\\ 
		\hline
	\end{tabular}
\end{table}

\subsection{Datasets and Experimental Design}

We want to evaluate the effectiveness and efficiency of the ComSR$_\text{non}$ and ComSR$_\text{ful}$ algorithms. This experiment uses a sign language utterance dataset (SIGN)\footnote{http://www.philippe-fournier-viger.com/spmf/index.php?link=datasets.php}. This is a real dataset, generated in real life. Each itemset has only one item. It is a sequence database of special items with sequential relationships between them. SIGN is a dense dataset, and the average length of each sequence is 51.997 (731 sequences, and 267 different items). Due to ComSR$_\text{non}$ having higher operating efficiency, this algorithm tests the entire SIGN dataset, while only the first 100 records of the SIGN dataset were taken for ComSR$_\text{ful}$ testing because of inefficiency. In addition, the thresholds of the former will be set lower accordingly, and the thresholds of the latter will be set higher to adapt to the hardware facilities. The threshold setting is shown in Table \ref{tab:exp_design}.

\subsection{Effectiveness Analysis}

First, we prove the effectiveness of the proposed algorithms. We illustrate some examples from the constantly updated sequential rules. We analyze the experimental results of complete compression of the ComSR$_\text{ful}$ algorithm while \textit{minsup} and \textit{minconf} are both 0.7. When the initial code set completely encodes the database, a total of 882 sequential rules (1 $\times$ 1) are used. In the experimental results of ComSR$_\text{ful}$, 876 sequential rules are finally obtained, including 4 $\times$ 1, 3 $\times$ 1, and 2 $\times$ 1 size of sequential rules, as well as 1 $\times$ 1 sequential rules, such as \{1, 18, 245, 17\} $\to$ \{253\}, \{1, 27245\} $\to$ \{253\}. Among them, \{1, 18, 245, 17\} $\to$ \{253\} is encoded 73 times, and \{1, 27245\} $\to$ \{253\} is encoded 15 times. According to the experimental results of ComSR$_\text{ful}$, part of the 1 $\times$ 1 size sequential rules are replaced by other sequential rules larger than 1 $\times$ 1 size for encoding, and the description length is also reduced. This shows that the number of sequential rules is reduced while the description of the database is maintained.

We can conclude from the preceding analysis that ComSR$_\text{non}$ and ComSR$_\text{ful}$ are trade-off coordination. ComSR$_\text{non}$ is suitable for finding only the sequential rule set that meets the thresholds. ComSR$_\text{ful}$ is more suitable for finding the sequential rule set that can fully describe the sequence database. In addition, the disadvantage of ComSR$_\text{non}$ will be ``over-compact'' for the sequential rule set. In other words, too few sequential rules are found, and the description level of the sequence database is low. The disadvantage of ComSR$_\text{ful}$ is that there are rules that do not meet the thresholds in the obtained sequential rule set.

\subsection{Number of Sequential Rules Analysis}

\begin{figure}[h]
	\centering
	\includegraphics[trim=50 0 0 0, scale=0.23]{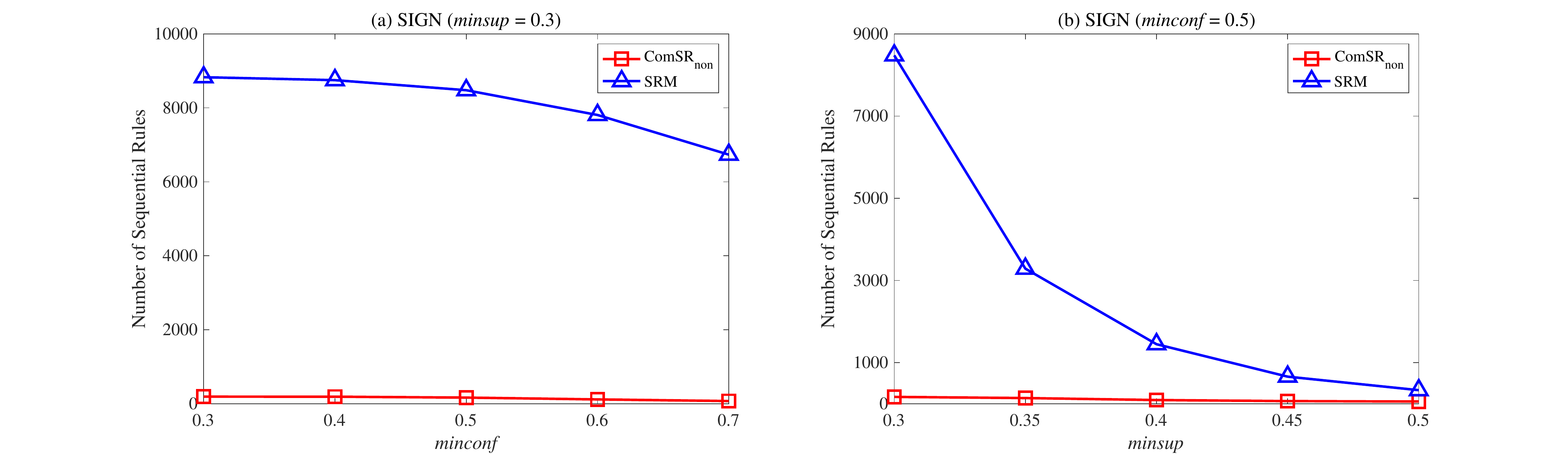}
	\caption{The number of sequential rules in ComSR$_\text{non}$ with various \textit{minsup} and \textit{minconf}.}
	\label{fig:ComSR_non_SR_size}
\end{figure}

In this subsection, the numbers of the two sets of sequential rules obtained by ComSR$_\text{non}$ or ComSR$_\text{ful}$ will be compared with each other. We also compare the number of sequential rules obtained by existing SRM algorithms, and analyze the effects on the compared algorithms. In Fig. \ref{fig:ComSR_non_SR_size}(a) (or Fig. \ref{fig:ComSR_non_SR_size}(b)), when \textit{minsup} (or \textit{minconf}) is fixed, the number of sequential rules in ComSR$_\text{non}$ gradually decreases as \textit{minconf} (or \textit{minsup}) increases. Considering Fig. \ref{fig:ComSR_non_SR_size}(a) and Fig. \ref{fig:ComSR_non_SR_size}(b), the obtained sequential rule sets are indeed very small, but they are not fully compressed. Therefore, we call this case as ``over-compact''. Even though these sequential rules can be used to describe the sequence database, it may not be a full enough description.

\begin{figure}[h]
	\centering
	\includegraphics[trim=50 0 0 0, scale=0.23]{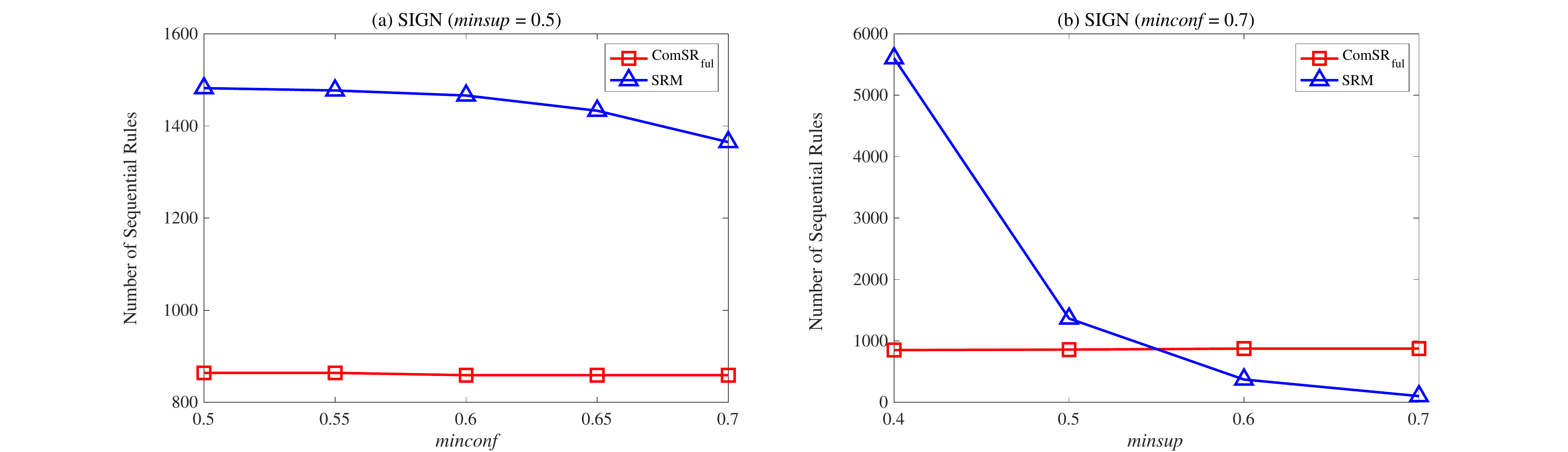}
	\caption{The number of sequential rules in ComSR$_\text{ful}$ with various \textit{minsup} and \textit{minconf}.}
	\label{fig:ComSR_ful_SR_size}
\end{figure}

The final set of sequential rules of ComSR$_\text{ful}$ will undoubtedly be larger. In Fig. \ref{fig:ComSR_ful_SR_size}(a), when the \textit{minsup} is fixed at 0.5, the overall number of sequential rules is also reduced while \textit{minconf} increases from 0.5 to 0.7. However, the number of sequential rules does not change greatly. The number of final sequential rules of ComSR$_\text{ful}$ is smaller than that of the existing sequential rule mining algorithm. This case shows that ComSR$_\text{ful}$ can find a set of more compact sequential rules under the condition of completely describing the data.

When the thresholds set by the user gradually decrease, the sequential rules mined by the SRM algorithm will increase continuously. In Fig. \ref{fig:ComSR_ful_SR_size}(b), when the \textit{minconf} is fixed, with the decrease of the \textit{minsup} from 0.7 to 0.4, the resulting sequential rule set keeps decreasing. Compared to the SRM algorithm, ComSR$_\text{ful}$ can not only describe the sequence database completely but also reduce the number of sequential rules. However, there are some sequential rules that do not reach the thresholds in the sequential rule set. Only the sequential rules that exceed the threshold values play the role of describing the characteristics of the sequence database. These sequential rules can describe the sequence database completely, but the characteristics of the sequence database are not very obvious.

\subsection{Compression Ratio Evaluation}

\begin{figure}[h]
	\centering
	\includegraphics[trim=50 0 0 0, scale=0.23]{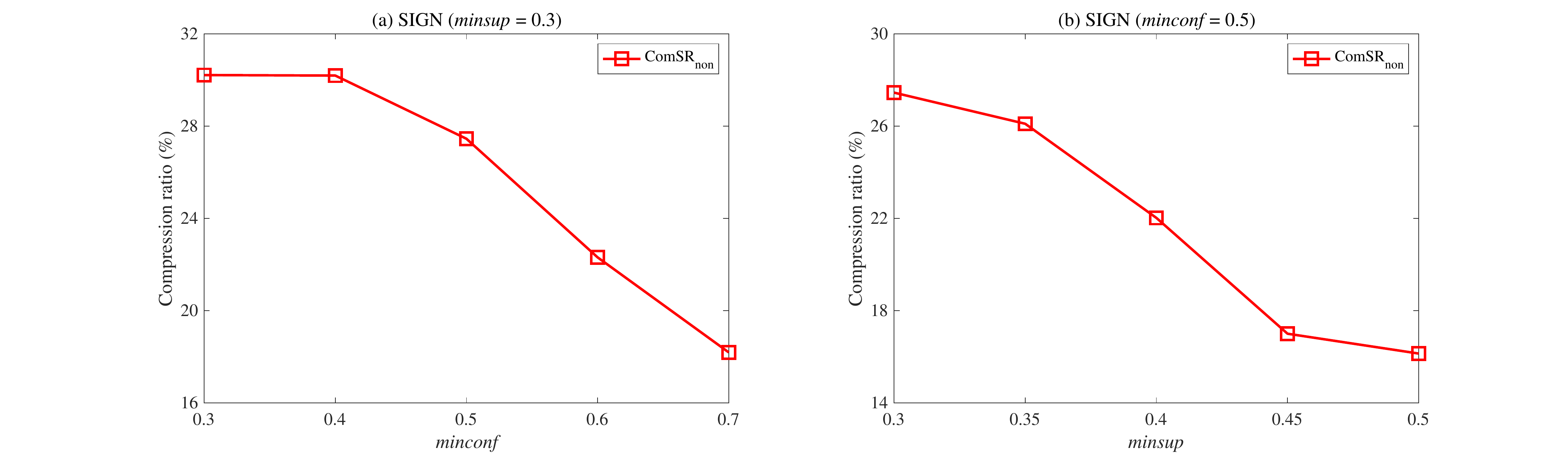}
	\caption{The compression ratio of ComSR$_\text{non}$ with various \textit{minsup} and \textit{minconf}.}
	\label{fig:ComSR_non_compression_ratio}
\end{figure}

In this subsection, we mainly analyze the ComSR$_\text{non}$ algorithm (as shown in Fig. \ref{fig:ComSR_non_compression_ratio}(a) and Fig. \ref{fig:ComSR_non_compression_ratio}(b)), and the ComSR$_\text{ful}$ algorithm can achieve 100\% complete compression for the database. The compression ratio is calculated as follows:

\begin{small}
    \begin{equation}
        \textit{compression ratio} = \frac{\sum_{seq \in \mathcal{D}}{|\{\textit{item}|\textit{item}  \in  \text{\textit{seq} if \textit{item} is encoded}\}|}}{\sum_{seq \in \mathcal{D}}{|\{\textit{item}|\textit{item} \in seq\}|}}.
    \end{equation}
\end{small}

In Fig. \ref{fig:ComSR_non_compression_ratio}(a), it can be seen that when \textit{minsup} is fixed at 0.3, the compression ratio decreases when \textit{minconf} increases. When \textit{minconf} is 0.3 or 0.4, the compression ratio does not change, but the compression ratio decreases more when \textit{minconf} is 0.5. This is because when \textit{minconf} is reduced from 0.4 to 0.3, there is no sequential rule that can better compress the sequence database. When the \textit{minconf} is reduced from 0.5 to 0.4, you can find the sequential rules for better compression of the sequence database. It is not difficult to find that even when the \textit{minsup} and the \textit{minconf} are set to 0.3 and 0.3, respectively, the compression ratio is still only 30\% $<$ 50\%. When \textit{minsup} and \textit{minconf} are set to 0.3 and 0.7, respectively, the compression ratio is even as small as 18.19\%. It can be said that the expected effect of compression has not been achieved. In Fig. \ref{fig:ComSR_non_compression_ratio}(b), when \textit{minconf} is 0.5, there is no doubt that the compression ratio for the sequence database is also continuously reduced if \textit{minsup} increases. Similarly, the compression ratio has not achieved the expected effect. For example, when the \textit{minsup} and the \textit{minconf} are set to 0.3 and 0.5, respectively, the compression ratio can only reach 27.45\%. In the above experimental analysis, the main reason for the low compression ratio is that the sequential rules mined by the existing SRM algorithms can only encode some items of some sequences in the database. When the threshold settings are infinitely close to 0, the sequential rules mined by the SRM algorithm must be able to completely encode the sequence database. However, there are high requirements for the performance of the existing SRM algorithms and the ComSR algorithm.

From the analysis of Fig. \ref{fig:ComSR_non_compression_ratio}(a) and Fig. \ref{fig:ComSR_non_compression_ratio}(b), we can learn that ComSR$_\text{non}$ does not perform very well on the compression of the sequence database. However, this does not imply that ComSR$_\text{non}$ cannot find a set of compact sequential rules at the end. The reason is that some sequential rules which do not meet the thresholds (e.g., \textit{minsup} and \textit{minconf}) cannot be added into the code set. Therefore, the items in these sequential rules cannot be encoded into the corresponding sequences. On the contrary, ComSR$_\text{ful}$ is very likely to return the sequential rules that do not meet the thresholds. Therefore, ComSR$_\text{non}$ and ComSR$_\text{ful}$ are trade-off coordination methods that can be used in different situations.

\subsection{Run Time Analysis}

The ComSR$_\text{non}$ algorithm's running time is then examined. It should be pointed out that the running time usage only calculates the part of looping to add the candidate sequential rule \textit{candidate} to code set \textit{CodeSet} (Lines 4--12 of Algorithm \ref{alg:ComSRnon} and Algorithm \ref{alg:ComSRful}). In the first part of the algorithm, the sequential rule set is found by using the existing sequential rule mining algorithm.

\begin{figure}[h]
	\centering
	\includegraphics[trim=50 0 0 0, scale=0.22]{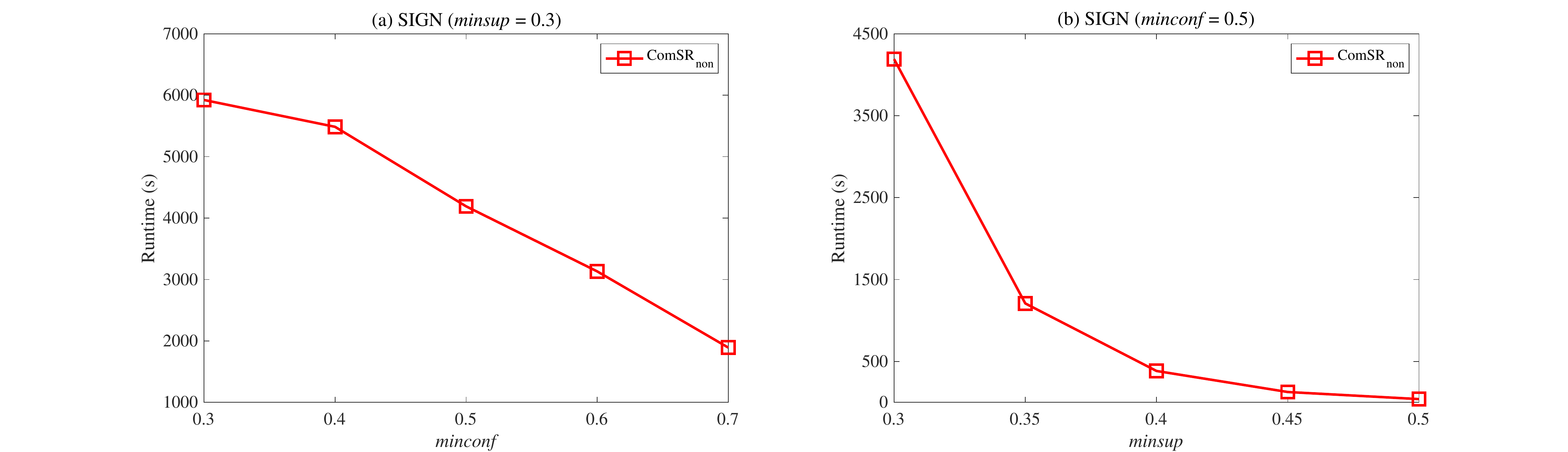}
	\caption{The running time of ComSR$_\text{non}$ with various \textit{minsup} and \textit{minconf}.}
	\label{fig:ComSR_non_runtime}
\end{figure}

In Fig. \ref{fig:ComSR_non_runtime}(a), when the \textit{minsup} is fixed at 0.3, the running time of ComSR$_\text{non}$ decreases as the \textit{minconf} increases from 0.3 to 0.7. In addition, the slope of Fig. \ref{fig:ComSR_non_runtime}(a) with \textit{minconf} from 0.3 to 0.4 is relatively gentle compared with the slope of 0.4 to 0.7. When the \textit{minconf} is fixed at 0.5 (Fig. \ref{fig:ComSR_non_runtime}(b)), it can be found that the running time decreases continuously as \textit{minsup} increases. However, when the \textit{minsup} decreases from 0.3 to 0.35, the slope of Fig. \ref{fig:ComSR_non_runtime}(b) is steeper than that of from 0.35 to 0.5. When the \textit{minsup} decreases from 0.35 to 0.5, the slope is relatively gentle. We can conclude that the running time decreases more and more slowly.

\begin{figure}[h]
	\centering
	\includegraphics[trim=50 0 0 0, scale=0.23]{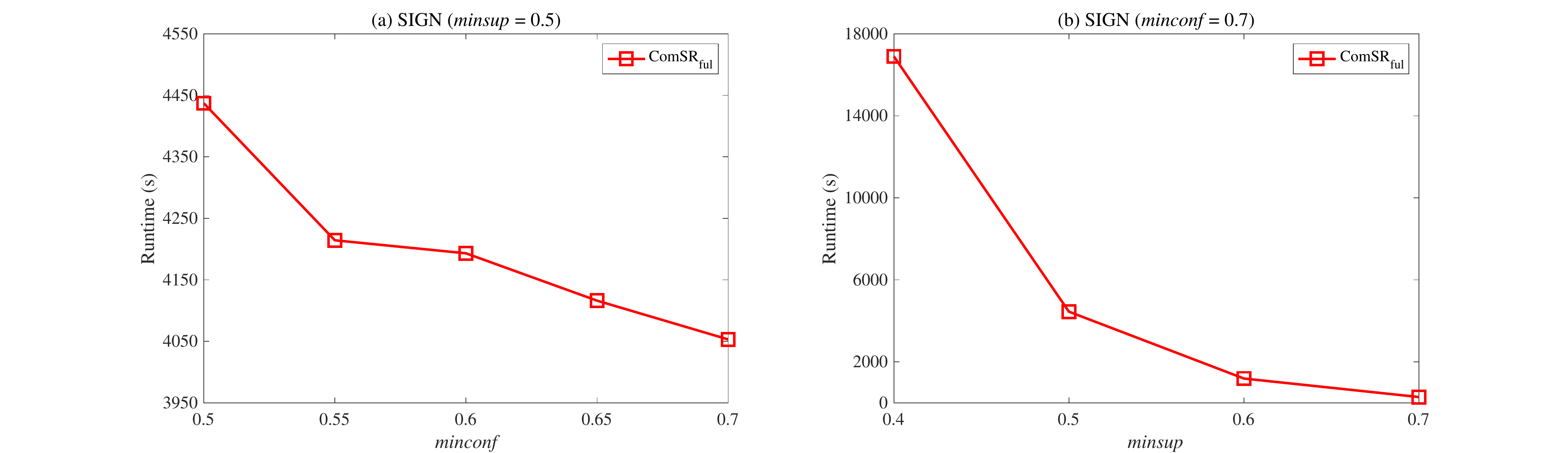}
	\caption{The running time of ComSR$_\text{ful}$ with various \textit{minsup} and \textit{minconf}.}
	\label{fig:ComSR_ful_runtime}
\end{figure}

Fig. \ref{fig:ComSR_ful_runtime} shows the runtime consumption of ComSR$_\text{ful}$. When the \textit{minsup} is fixed at 0.5, the \textit{minconf} increases from 0.5 to 0.7, the running time also decreases. When the \textit{minconf} is fixed at 0.7 and then the \textit{minsup} is changed, we can find that the running time is gradually reduced. However, the reduction is very large. This situation shows that when \textit{minconf} is fixed to 0.7 and \textit{minsup} is reduced from 0.5 to 0.4, more sequential rules are generated, and thus more sequential rules are used as the code set, resulting in an increase in the running time. When the \textit{minsup} is 0.4, the running time is nearly 4x, 14x, and 59x faster than if \textit{minsup} is 0.5, 0.6, and 0.7, respectively. The gap in running time under different \textit{minsup} degrees is large.

Through the comparison of Fig. \ref{fig:ComSR_ful_runtime}(a) and Fig. \ref{fig:ComSR_ful_runtime}(b), when the \textit{minconf} is fixed, the reduction range of the running time of changing the \textit{minsup} is greater than that of fixing the \textit{minsup} and changing the \textit{minconf}.

\section{Conclusion} \label{sec:conclusion}

In a large-scale sequence database, when the given thresholds \textit{minsup} and \textit{minconf} become small, the number of mined sequential rules may increase explosively. Thus, it is difficult to find the regular characteristics of the sequence database with a sequential relationship. In this paper, we propose a heuristic algorithm that combines the Minimum Description Length principle with the SRM algorithm to find a set of compact and meaningful sequential rules. We propose two algorithms, ComSR$_\text{non}$ and ComSR$_\text{ful}$, which are balanced and coordinated choices. The conducted experiments on the SIGN dataset prove that the novel algorithms are effective and feasible. In future work, targeted improvements can be made according to the different shortcomings of different algorithms. The compression ratio of ComSR$_\text{non}$ can be improved. For example, the coding order of different sequential rules can be adjusted. The time complexity of ComSR$_\text{ful}$ can be reduced, and its basic code set can be appropriately filtered. In addition, the algorithms proposed in this paper have some special requirements for the dataset of sequential rules. As a result, they can be improved to adapt to different types of datasets.

\ifCLASSOPTIONcaptionsoff
  \newpage
\fi

\bibliographystyle{IEEEtran}
\bibliography{ComSR}

\end{document}